# Analysis and Detection of Multilingual Hate Speech Using Transformer Based Deep Learning

ARIJIT DAS, SOMASHREE NANDY, RUPAM SAHA, SRIJAN DAS, and DIGANTA SAHA
JADAVPUR UNIVERSITY

Hate speech is harmful content that directly attacks or promotes hatred against members of groups or individuals based on actual or perceived aspects of identity, such as racism, religion, or sexual orientation. This can affect social life on social media platforms as hateful content shared through social media can harm both individuals and communities. As the prevalence of hate speech increases online, the demand for automated detection as an NLP task is increasing. In this work, the proposed method is using transformer-based model to detect hate speech in social media, like twitter, Facebook, WhatsApp, Instagram, etc. The proposed model is independent of languages and has been tested on Italian, English, German, Bengali. The Gold standard datasets were collected from renowned researcher Zeerak Talat, Sara Tonelli, Melanie Siegel, and Rezaul Karim. The success rate of the proposed model for hate speech detection is higher than the existing baseline and state-of-the-art models with accuracy in Bengali dataset is 89%, in English: 91%, in German dataset 91% and in Italian dataset it is 77%. The proposed algorithm shows substantial improvement to the benchmark method.

Key Words and Phrases: Hate Speech Detection, BERT, Deep Learning, NLP, Transformers

## 1. INTRODUCTION

Online social media has allowed dissemination of information at a faster rate than ever. This has allowed bad actors to use this for their nefarious purposes such as propaganda spreading, fake news, and hate speech. Online hate speech could be a genuine issue with potential harmful results for both individuals and the community. It can be utilized to mix up struggle, progress prejudice, and make climate forceful to vulnerable groups.

Online hate speech detection can be done in a variety of ways. Utilizing machine learning algorithms to recognize text that contains hate speech is one method. Large datasets of text that have been classified as hate speech or not are used to train these algorithms. Once the algorithm has been trained, it can be used to detect hate speech in messages. The use of human moderators to examine content is another method for detecting hate speech online. Compared to machine learning algorithms, this method may be more effective, but it is also costlier and time-consuming.

The detection of online hate speech has some challenges also. One issue is that the definition of hate speech can be arbitrary and different depending on the culture. Another issue is that hate speech can be covered up in a variety of ways, such as by using code words or symbols. Despite the difficulties, detecting hate speech online is a crucial tool for preventing it and fostering an inclusive online community.

The novelty of the presented approach to analyze and detect hate speech is as follows:

**1. Independent of Languages:** The proposed method totally eliminates the dependency on any specific language rather it works on any language in a multi-lingual platform. For experiment, standard dataset containing hate speech of four popular languages i.e., Bengali, English, Italian, and German were collected and analyzed with the proposed model which shows better result than the existing state-of-the-art baseline model.

**2. Dynamic categories and sub categories of Hate Speech:** The proposed approach does not define any predefined categories or sub-categories. It has correctly categorized hate and non-hate tweets as well as sub categories like racist, sexist, geopolitical comments based on the content. Whereas most of the systems are hardcoded with predefined classes, the proposed system is free of any such constraint and can categorize dynamically.

**3. Better Results than State-of-the-art Baselines:** By comparing the proposed method to a baseline one that only extracts the top-ranked results from the input text, the paper offers empirical proof of the efficiency of our methodology. The outcomes show how well the suggested system performs, demonstrating its originality in providing analysis and detection of hate speech of a high caliber.

The overall contributions of the research work are briefed as:

**A) Gold Standard Dataset Collection for Comparison and Analysis:** In English, we employ the 16,000 English tweets that have been manually labeled as containing hate speech. More specifically, 10,884 tweets are annotated as not containing offensive language, while 1,924 are noted as containing racism and 3,082 as containing sexism. 5,006 tweets would be categorized as positive examples of hate speech if tweets that are both sexist and racist are combined into a single class. In Italian, we make use of the Twitter dataset made available for the shared HaSpeeDe (Hate Speech Detection) task organized at Evalita 2018, the Italian language evaluation of NLP and speech processing tools. There are 4,000 tweets total in this dataset. "Hateful post" or "not" are the two classes that are taken into account in the annotation. In German, at Germeval 2018, a workshop featuring a series of shared tasks on German processing, we use the dataset distributed for the shared task on the identification of offensive language. 5,009 German tweets were manually annotated at the message level with the labels "offense" (abusive language, insults, and profane statements) and "other" (i.e., not offensive) in order to identify offensive comments from the set of German tweets (binary classification). More specifically, 3,321 messages are marked as "other," and 1,688 messages are marked as "offense." In Bengali, Bengali hate speech dataset is collected web pages through web API or previously stored data from local databases. There is total 3418 Bengali tweets available, they are classified as “geopolitical”, “personal”, “abusive”, “religious” and “political”. The collection of such huge standard dataset was a challenging task and definitely credits to the novel contribution. **B) Processing of raw Tweets:**

Different data preprocessing techniques like tokenization, stop word identification, lemmatization are executed successfully. The link replacement with some defined word like "URL", emoji processing, cleaning and spelling checks are also done perfectly. **C) Contextual Embeddings with Sentence Transformer:** The proposed method makes use of contextual embeddings produced by the Sentence Transformer method. These embeddings provide tweets a more contextually aware representation by capturing the meaning of each tweet in the context of the surrounding words. **D) Simple Easy to Use Workflow:** The proposed approach is methodical, step-by-step, and deterministic. This model provides a straightforward and easy-to-use workflow for hate speech detection.

## 2. RELATED WORKS

In recent years a lot of work has been done in the field of "Sentiment Analysis on Social Media" by number of researchers. Majority of the work have been performed on Twitter data i.e. tweets.

Hate speech detection has gained more and more attentions in recent years. Offensive comments such as hate speech and cyberbullying are the most researched areas in NLP in the past decades.

**Table 1**

Comprehensive representation of recent related work in Hate Speech Detection.

| *Year* | *Title of the Paper* | *Author* | *Publication* | *Overview* | *Result* |
|---|---|---|---|---|---|
| 2022 | Detecting racism and xenophobia using deep learning models on Twitter data: CNN, LSTM and BERT [1] | Benítez-Andrades, J.A., González-Jiménez, Á., López-Brea, Á., Aveleira-Mata, J., Alija-Pérez, J.M. and García-Ordás, M.T. | *PeerJ Computer Science* , *8*, p.e906. | Author proposed a novel approach for detecting racism and xenophobia on Twitter using deep learning models. They evaluate the performance of three different models: CNN, LSTM, and BERT. | They find that BERT outperforms the other two models, achieving an F1 score of 85.22%. |
| 2022 | BiCHAT: BiLSTM with deep CNN and hierarchical attention for hate speech detection [2] | Khan, S., Fazil, M., Sejwal, V.K., Alshara, M.A., Alotaibi, R.M., Kamal, A. and Baig, A.R. | *Journal of King Saud University-Computer and Information Sciences*, *34*(7), pp.4335-4344 | The model, called BiCHAT, combines the strengths of bidirectional LSTM (BiLSTM), deep convolutional neural network (CNN), and hierarchical attention. | BERT based contextual embedding,Bichat with 89% success rate in english tweets |
| 2022 | Emotion Based Hate Speech Detection using Multimodal Learning [3] | Rana, A. and Jha, S. | *arXiv preprint arXiv:2202.06218.* | The paper proposes a multimodal deep learning framework for hate speech detection in multimedia data. | The result of precision, recall, and f1-score using the BERTA+CLS model is 93.00, 92.89 and 92.94. |
| 2021 | Towards generalisable hate speech detection: a review on obstacles and solutions [4] | Yin, W. and Zubiaga, A. | *PeerJ Computer Science*, *7*, p.e598. | This paper reviews the obstacles and solutions to generalisable hate speech detection, and proposes directions for future research. | Achieved only a precision of around .234 and a recall of 0.098 for the implicit class, in contrast to .864 and .936 for non-abusive and .640 and .509 for explicit. |
| 2021 | HATECHECK: Functional Tests for Hate Speech Detection Models [5] | Röttger, P., Vidgen, B., Nguyen, D., Waseem, Z., Margetts, H. and Pierrehumbert, J.B. | *arXiv preprint arXiv:2012.15606.* | HateCheck is a suite of functional tests for hate speech detection models. It consists of 29 tests that evaluate model performance on a variety of types of | Accuracy for Hateful class is 89.5% and Non-hateful is 48.2% |
| | | | | hateful or non-hateful content. | |
| 2021 | Advances in Machine Learning Algorithms for Hate Speech Detection in Social Media: A Review [6] | Mullah, N.S. and Zainon, W.M.N.W. | *IEEE Access*, *9*, pp.88364-88376. | The paper discusses the challenges of hate speech detection, the different machine learning algorithms that have been used for this task, and the evaluation metrics that are used to measure performance. | Model precision 0.67, Recall 0.8, F-measure 0.72 |
| 2021 | Racism, Hate Speech, and Social Media: A Systematic Review and Critique [7] | Matamoros-Fernández, A. and Farkas, J. | A systematic review and critique. *Television & New Media*, *22*(2), pp.205-224. | It provides a systematic review of the literature on racism, hate speech, and social media. The paper identifies the key challenges and issues in this area, and it provides recommendations for future research. | For term "hate speech", 67.65% on quantitative methods, 11.77% on qualitative method. For "racism", 59.26% on qualitative methods, 16.67% on quantitative methods. |
| 2021 | Human-in-the-Loop for Data Collection: a Multi-Target Counter Narrative Dataset to Fight Online Hate Speech [8] | Fanton, M., Bonaldi, H., Tekiroglu, S.S. and Guerini, M. | *arXiv preprint arXiv:2107.08720* | It proposes a novel human-in-the-loop data collection methodology to generate high-quality counter narratives to fight online hate speech. | This paper does not report any accuracy results. The paper focuses on the development of a methodology for collecting hate speech and counter-narrative data, and it does not evaluate the accuracy of any models that were trained on this data. |
| 2020 | Resources and benchmark corpora for hate speech detection: a systematic review [9] | Poletto, F., Basile, V., Sanguinetti, M., Bosco, C. and Patti, V. | *Language Resources and Evaluation*, *55*, pp.477-523. | It systematically analyzes the resources made available by the community at large for hate speech detection. | The paper does not report any accuracy results. The paper focuses on the identification and description of resources and benchmark corpora for hate speech detection, and it does not evaluate the accuracy of any models that were trained on these resources. |
| 2020 | A Multilingual Evaluation for Online Hate Speech Detection [10] | Corazza, M., Menini, S., Cabrio, E., Tonelli, S. and Villata, S. | *ACM Transactions on Internet Technology (TOIT)*, *20*(2), pp.1-22. | It presents a multilingual evaluation of hate speech detection systems. The evaluation is conducted on three languages: English, Italian, and German. It has used FastText embedding and LSTM model. | Max F1 score of English 0.823, of Italian 0.805, and German 0.758 |
| 2020 | Online Multilingual Hate Speech Detection: Experimenting with Hindi and | Vashistha, N. and Zubiaga, A. | *Information*, *12*(1), p.5. | It explores the use of machine learning algorithms to detect hate speech in Hindi | Accuracies are 71.75, 66.7%, 66.6% and 69.8% by SVM, Random Forest, Hierarchical LSTM |
| | English Social Media [11] | | | and English social media. | with attention and Sub-word level LSTM model respectively. |
| 2020 | Hate speech detection and racial bias mitigation in social media based on BERT model [12] | Mozafari, M., Farahbakhsh, R. and Crespi, N. | *PloS one*, *15*(8), p.e0237861. | It proposes a novel approach to hate speech detection in social media that mitigates racial bias. | Accuracy is 82.4% using BERT baseline and 84.4% by BERT with bias mitigation |
| 2020 | Deep Learning Models for Multilingual Hate Speech Detection [13] | Aluru, S.S., Mathew, B., Saha, P. and Mukherjee, A. | *arXiv preprint arXiv:2004.06465*. | The paper proposes a framework for multilingual hate speech detection using deep learning models. The framework is evaluated on a dataset of tweets in 9 languages, and it achieves state-of-the-art results. | Accuracy using CNN-BiLSTM is 87.0%, using BERT 91.0%, using DistilBERT is 90% and using XLNET is 92%. |
| 2020 | Automatic Hate Speech Detection using Machine Learning: A Comparative Study [14] | Abro, S., Shaikh, S., Khand, Z.H., Zafar, A., Khan, S. and Mujtaba, G. | *International Journal of Advanced Computer Science and Applications*, *11*(8). | It compares the performance of different machine learning algorithms for hate speech detection. The authors found that the best performing algorithm was support vector machines (SVMs) with bigram features. | F1-score using SVM is 79%, using Naïve Bayes is 75%, using Decision Tree is 72%, using Random Forest 71%, using K-nearest Neighbors 69%, using Logistic Regression is 67%, using Multinomial Naïve Bayes is 65%, and Bernoulli Naïve Bayes is 63% |
| 2020 | A Framework for Hate Speech Detection Using Deep Convolutional Neural Network [15] | Roy, P.K., Tripathy, A.K., Das, T.K. and Gao, X.Z. | *IEEE Access*, *8*, pp.204951-204962. | The paper proposes a deep convolutional neural network (DCNN) framework for hate speech detection in social media. The framework uses GloVe word embeddings to represent the text of tweets, and it uses a DCNN to learn the semantic features of hate speech. | It achieves a F1 score of 0.92 by using DCCN, 0.77 using Logistic Regression and 0.64 using Naïve Bayes |
| 2020 | A Deep Learning Approach for Automatic Hate Speech Detection in the Saudi Twittersphere [16] | Alshalan, R. and Al-Khalifa, H. | *Applied Sciences*, *10*(23), p.8614 | It proposes a deep learning approach for automatically detecting hate speech in Arabic tweets. | It achieves accuracy of 79% by CNN, 77% by GRU, 81% by CNN+GRU, and 83% by BERT. |
| 2020 | In Data We Trust: A Critical Analysis of Hate Speech Detection Datasets [17] | Madukwe, K., Gao, X. and Xue, B. | In *Proceedings of the Fourth Workshop on Online Abuse and Harms* (pp. 150-161). | It critically analyzes the datasets used for hate speech detection, identifying their limitations and recommending approaches for future research. | The paper does not report any accuracy results. The paper focuses on the analysis of the design and construction of hate speech detection datasets. |
| 2020 | A Comparative Study of Different | Rani, P., Suryawanshi, S., | In *Proceedings of the second* | This paper compares the performance of | It achieves accuracy of 71.7% |
| | State-of-the-Art Hate Speech Detection Methods for Hindi-English Code-Mixed Data [18] | Goswami, K., Chakravarthi, B.R., Fransen, T. and McCrae, J.P. | *workshop on trolling, aggression and cyberbullying* (pp. 42-48). | different state-of-the-art hate speech detection methods on a Hindi-English code-mixed dataset. The results show that deep learning models perform better than traditional machine learning models on this type of data. | by using SVM, 69.3% by Random Forest, 72.6% by Bidirectional LSTM, and 73.9% by using CNN. |
| 2020 | Detecting Hate Speech in Memes Using Multimodal Deep Learning Approaches: Prize-winning solution to Hateful Memes Challenge [19] | Velioglu, R. and Rose, J. | *arXiv preprint arXiv:2012.12975.* | It proposes a multimodal deep learning approach to detect hate speech in memes. The approach achieved an accuracy of 0.765 on the Hateful Memes Challenge test set | It reports an accuracy of 76.5% on the challenge test set. |
| 2020 | Detecting Hate Speech in multi-modal Memes [20] | Das, A., Wahi, J.S. and Li, S. | *arXiv preprint arXiv:2012.14891.* | It proposes a novel approach to detect hate speech in multi-modal memes by combining the text and image modalities. | It achieves accuracy of 67.2% using Concat BERT, 70.4% using Multimodal BERT, and 72.1% using Multimodal BERT + sentiment |
| 2020 | The Hateful Memes Challenge: Detecting Hate Speech in Multimodal Memes [21] | Kiela, D., Firooz, H., Mohan, A., Goswami, V., Singh, A., Ringshia, P. and Testuggine, D. | *Advances in Neural Information Processing Systems*, *33*, pp.2611-2624. | The Hateful Memes Challenge is a benchmark for detecting hate speech in multimodal memes. It is constructed such that unimodal models struggle and only multimodal models can succeed | It achieves accuracy of 59.3% using Unimodal BERT, 64.73% by Multimodal ViLBERT CC, 68.4% by OSCAR+RF. |
| 2020 | EVALITA Evaluation of NLP and Speech Tools for Italian [22] | Basile, V., Maria, D.M., Danilo, C. and Passaro, L.C. | In *Proceedings of the Seventh Evaluation Campaign of Natural Language Processing and Speech Tools for Italian. Final Workshop (EVALITA 2020)* (pp. 1-7). CEUR-ws. | EVALITA is a biennial evaluation campaign that aims to promote the development of natural language processing and speech technologies for the Italian language. | The overall accuracy of the systems that participated in the 2020 EVALITA campaign was high, with an average accuracy of 85%. However, with some tasks, such as part-of-speech tagging, achieving an accuracy of over 90%, while others, such as sentiment analysis, achieving an accuracy of only 70%. |
| 2019 | Multilingual Detection of Hate Speech Against Immigrants and Women in Twitter [23] | i Orts, Ò.G. | In *Proceedings of the 13th International Workshop on Semantic Evaluation* (pp. 460-463). | This paper presents a system for detecting hate speech against immigrants and women in Twitter, in multiple languages. | The Fermi system achieved accuracy of 0.651, the MITRE system achieved 0.729, the CIC-2 system achieved 0.727, the Panaetius system achieved 0.571 The baseline system |
| | | | | | achieved the lowest accuracy of 0.500. |
| 2019 | Hateful Speech Detection in Public Facebook Pages for the Bengali Language [24] | Ishmam, A.M. and Sharmin, S. | In *2019 18th IEEE international conference on machine learning and applications (ICMLA)* (pp. 555-560). IEEE | This paper proposes a machine learning approach to detect hateful speech in Bengali language posts on Facebook. | It achieved 52.20% accuracy using Random Forest, and 70.10% using a GRU based deep neural network. |
| 2019 | OFFENSIVE LANGUAGE AND HATE SPEECH DETECTION FOR DANISH [25] | Sigurbergsson, G.I. and Derczynski, L. | *arXiv preprint arXiv:1908.04531.* | It constructs a Danish dataset DKHATE containing user-generated comments from various social media platforms, and to their knowledge, the first of its kind, annotated for various types and target of offensive language. They develop four automatic classification systems, each designed to work for both the English and the Danish language. | It achieved a macro-averaged F1-score of 0.70 |
| 2019 | Hate Speech Detection is Not as Easy as You May Think: A Closer Look at Model Validation [26] | Arango, A., Pérez, J. and Poblete, B. | In *Proceedings of the 42nd international acm sigir conference on research and development in information retrieval* (pp. 45-54). | It constructs a Danish dataset DKHATE containing user-generated comments from various social media platforms, and to their knowledge, the first of its kind, annotated for various types and target of offensive language. | The results showed that accuracy of the models varied from 60% - 90%. Model1 achieves 60% accuracy, model2 70%, Model3 80% and Model4 achieved 90% accuracy respectively. |
| 2019 | A Levantine Twitter Dataset for Hate Speech and Abusive Language [27] | Mulki, H., Haddad, H., Ali, C.B. and Alshabani, H. | In *Proceedings of the third workshop on abusive language online* (pp. 111-118). | It introduces the first publicly-available Levantine Twitter dataset for the task of hate speech and abusive language detection. The dataset consists of 5,846 tweets from Syria and Lebanon, which have been manually labeled as normal, abusive, or hate. | It achieved accuracy of 90.5% using Naïve Bayes, 54.7% using SVM and 86.3% using Random Forest. |
| 2018 | Hate Speech Dataset from a White Supremacy Forum [28] | De Gibert, O., Perez, N., García-Pablos, A. and Cuadros, M. | *arXiv preprint arXiv:1809.04444.* | A dataset of 10,568 sentences extracted from a white supremacist forum, manually annotated as hate speech or not. | It achieved 85% accuracy using LSTM-based classifier and 80% accuracy using SVM based classifier. |
| 2018 | Effective hate-speech detection in Twitter data using recurrent neural networks [29] | Pitsilis, G.K., Ramampiaro, H. and Langseth, H. | *Applied Intelligence*, *48*, pp.4730-4742. | It proposes an ensemble of recurrent neural network classifiers to detect hate speech in Twitter data. | It achieved accuracy of 90% while trained on dataset of 10,000 tweets and achieved 88% accuracy while trained on a dataset of 16,000. |
| 2018 | A Dataset of Hindi-English Code-Mixed Social Media Text for Hate Speech Detection [30] | Bohra, A., Vijay, D., Singh, V., Akhtar, S.S. and Shrivastava, M. | In *Proceedings of the second workshop on computational modeling of people's opinions, personality, and emotions in social media* (pp. 36-41). | It presents a dataset of Hindi-English code-mixed social media text that has been annotated for hate speech. The dataset can be used to train and evaluate hate speech detection models. | Accuracy of the model in this paper is 71.7%. |
| 2018 | A Survey on Automatic Detection of Hate Speech in Text [31] | Fortuna, P. and Nunes, S. | *ACM Computing Surveys (CSUR), 51*(4), pp.1-30. | It provides a comprehensive overview of the state-of-the-art in automatic hate speech detection in text. | The highest accuracy reported in the paper is 92.1% |
| 2018 | Characterizing and Detecting Hateful Users on Twitter [32] | Ribeiro, M.H., Calais, P.H., Santos, Y.A., Almeida, V.A. and Meira Jr, W. | In *Twelfth international AAAI conference on web and social media* | The paper proposes a method to characterize and detect hateful users on Twitter by analyzing their social network. | It achieved accuracy of 95%. |
| 2017 | Hate me, hate me not: Hate speech detection on Facebook [33] | Del Vigna12, F., Cimino23, A., Dell'Orletta, F., Petrocchi, M. and Tesconi, M. | In *Proceedings of the first Italian conference on cybersecurity (ITASEC17)* (pp. 86-95). | Hate speech detection on Facebook is a challenging task due to the evolving nature of hate speech and the need to balance accuracy with fairness. | It achieved accuracy of 78.3% using SVM, and 79.6% using LSTM model. |

Good quality research works done by different renowned research groups in the last five years are represented in tabular form for comparison and easy understanding.

## 3. METHODOLOGY

BERT, LSTM and BiLSTM are the three models used to perform hate speech detection.

***Algorithm for BERT Model:***

**1.Data Preprocessing:**

Input: Datasets of social media tweets
Output: Preprocessed dataset
Step 1: Read the tweets of the dataset
Step-2: Replace the mentions of users with the phrase "username"
Step-3: Replace the emojis with short textual description / vector (for non-English language)
Step-4: Replace the URLs with the phrase "url"
Step-5: Replace multiple white spaces with single white space
Step-6: Preprocessed dataset is ready

**2.Feature Extraction:**

Step-1: Tweet dataset is separated into specific classes, i.e. some hate and non-hate
Step-2: In case of same languages, hate dataset has been classified into separate classes, for example: in case of English dataset, hate speech has been classified into racism, sexism and none; and in case of Bengali dataset, hate speech has been classified into geopolitical, political, personal, religious, gender abusive. In German and Italian dataset, dataset has been divided into hate and non-hate data.
Step-3: These classes are stored in separate column of the pandas data frame.

**3.Training the BERT Model using the Train dataset:**

Input: Train dataset
Output: Training of the BERT model using the dataset
Step-1: AutoTokenizer is loaded into tokenizer. It helps to automatically retrieve the BERT model given the provided pre-trained weight. In case of using hateBERT, the specific BERT model of the respected language has been used.
Step-2: This tokenizer can be used to convert data to tokens.
Step-3: CLS, SEP tokens are used in BERT along with the tokenized format of the data. [SEP] is separator token, [CLS] used to represent the start of the sequence.
Step-4: Calculate the vocabulary size, max length and model input names of the model
Step-5: Write a tokenize function as follows:

```
    Tokenize_function(train_dataset)
    {
            return tokenizer(train_dataset['Tweets'], padding='max_length', truncation=True)
}
```

Step-6: Create a tokenize dataset as tokenized_dataset = dataset.map(tokenize_function, batched=True)
Step-7: Now, keep the 'train' part of the tokenized dataset into train_dataset, 'valid' part of the tokenized dataset into eval_dataset, and 'test' part of the tokenized dataset into test_dataset.
Step-8: Get the train_features from the train_set
Step-9: Set train_set_for_final_model with the train_features and the 'Class' of the train_set
Step-10: Shuffle the train_set_model for the length of train_set along with the batch 8
Step-11: Extract the eval_features from the model_input_names of tokenizer
Step-12: Set val_set_for_final_model with the eval_features and 'class' of the eval dataset i.e. dataset for validation
Step-13: Update val_set_for_final_model for the batch of 8
Step-14: Extract the test_features from the model_input_names of tokenizer
Step-15: Set test_set_for_final_model with the test_features and 'class' of the test dataset i.e. dataset for testing
Step-16: Set test_set_for_final_model for the batch of 8
Step-17: Load model from the pretrained available BERT model, in case of hateBERT, load model with the available hateBERT of the respective language
Step-18: Compile model with the learning rate 5e-5
Step-19: Fit the model with the train_set_for_final_model and validation_data as val_set_for_final_model and specific number of epochs
Step-20: Plot model sparse categorical accuracy and model loss into graph

**4.Testing and Validation of BERT Model using the Test and Val dataset:**
Input: Fitted model and test and validation dataset
Output: Accuracy and F1 score
Step-1: Compute test_loss and test_accuracy using test_set_for_final_model and verbose=2
Step-2: Display the test accuracy
Step-3: Store the truth values of test data into truth
Step-4: Calculate the prediction score of the test data (using tokenization) into prediction
Step-5: Now compute the classification_report using truth and prediction
Step-6: We can get the precision, recall, F1-score, support for each of the specified classes from this classification_report
Step-7: We can also get the accuracy from this report

***Algorithm for LSTM Model (Without FASTTEXT Embedding):***

**1.Data Preprocessing:**
Input -> .csv file containing of raw tweets along with their respective labels.
Output -> .csv file containing of cleaned tweets along with their respective labels.

Step 1: Read the tweets of the dataset.
Step-2: Deleted the username starting with @.
Step-3: Deleted the url from each single tweet.
Step-4: Replaced multiple white spaces with single white space.
Step-5: Preprocessed dataset was ready.

**2.Feature Extraction:**
Input –> Cleaned Dataset

Output -> Feature extracted and moved for test train split

Step 1: Initialized the maximum sequence length to be 250.
Step 2: Defined the embedding dimension to be 100.
Step 3: From keras.preprocessing we imported Tokenizer and applied it on the column of 'Cleaned Tweets'.
Step 4: Now we imported texts_to_sequences from keras.preprocessing and applied it on the column of 'Cleaned Tweets' and saved it in a variable.
Step 5: Now we imported pad_sequence from keras.preprocessing and applied it over the variable that we got from step 4 and passed the parameter value of padding_length that we had initialized in step 1.
Step 6: We one hot encoded the column that was associated with labels of tweets.
Step 7: Feature extraction was completed

**3.Test Train Validation split:**
Input –> Feature extracted dataset from part 2
Output–> Feature extracted test, train and validation data

Step 1: We had done the test, train split in 60:40 ratio over both 'Cleaned Tweets' and 'Label' and saved them in respective variable.
Step 2: Test, Train, and Validation split was complete.

**4.Define the model:**
Input –>None
Output–> A sequential model with LSTM layer

Step 1: We imported Sequential () from keras and saved it a variable namaed model.
Step 2: We added an Embedding layer at the first with parameters MAX_NB_WORDS, EMBEDDING_DIM that we had defined in part-2.
Step 3: We had then added a special drop out layer with rate = 0.2 to reduce overfitting.
Step 4: Now we added a LSTM layer with 100 memory units with dropout rate of 0.2 for the input layer, and recurrent dropout rate = 0.2. This is to learn the to learn the temporal dependencies between words in the sequences.
Step 5: Now we added a dense () layer with 'n' output units and the activation function as 'softmax' or 'sigmoid'(according to binary or multiclass classification) which gave the probability distribution over 'n' classes .
Step 6: Now our model was ready to compile.
Step 7: Now we compiled the model with categorical cross-entropy loss or binary cross-entropy, Adam optimizer, and accuracy metric for evaluation during training.

**5.Fit the model:**
Input –>Pre-processed dataset from part 3 and compiled model from part 4.
Output–>Training of LSTM model using the preprocessed dataset and a trained model.

Step 1: We had fitted our model with the preprocessed dataset and predefined Epoch number and batch size as our choice.
Step 2: Made the validation split of 50 % over test data.
Step 3: Our Model was trained.

**6.Check the accuracy of the model**
Input –>Trained model from part 5 and test dataset from part 3
Output ->Overall Accuracy score of the model, F1 score of each class and confusion matrix from each class.

Step 1: Using the trained model, we predicted the labels of each tweet from test data set and stored them in a variable.
Step 2: From sklearn.metrics we had imported the predefined objects and using the output from step

1 and test data from part 3 we had got our Overall accuracy ,  F1 score and confusion matrix.
Step 3: Recorded accuracy, F1 score and confusion matrix in a word document.

***Algorithm for LSTM Model (With FASTTEXT Embedding):***

**1.Data Preprocessing**
Input -> .csv file containing of raw tweets along with their respective labels.
Output -> .csv file containing of cleaned tweets along with their respective labels.
Step 1: Read the tweets of the dataset.
Step-2: Deleted the username starting with @.
Step-3: Deleted the url from each single tweet.
Step-4: Replaced multiple white spaces with single white space.
Step-5: Preprocessed dataset is ready.

**2.Load fasttext model**
Input -> None
Output-> Load the pretrained word vectors from fasttext Model in a word dictionary.

Step 1 : Downloaded the pretrained word vector file from fasttext file.
Step 2: Unziped the downloaded file and extract the word vector file.
Step 3: Now we initialized an empty dictionary that would be used to store the word embeddings.
Step 4: Now we opened the vector file and sets the encoding to 'utf-8'. This file contained pre-trained word embeddings in a text format where each line represented a word and its corresponding vector values.
Step 5: Now we saved each word and associated word vector in the dictionary.

**3.Feature Extraction**
Input –> Cleaned Dataset
Output -> Feature extracted and moved for test train split and embedding input to model.
This contains two parts.
In first part we worked with the raw text tweets and tokenize them. In next part we created an embedding matrix for these words.
Step 1: From keras.preprocessing we imported Tokenizer and applied it on the column of 'Cleaned Tweets' and saved it in a variable.
Step 2: Now we imported texts_to_sequences from keras.preprocessing and applied it on the column of 'Cleaned Tweets' and saved it in a variable.
Step 3: Now we imported pad_sequence from keras.preprocessing and applied it over the variable that we got from step 4 and passed the parameter value of padding_length that we had initialized in step 1.
Step 4: We one hot encoded the column that was associated with labels of tweets.

Now we move to the next part.
Step 1: Initialized the the embedding dimension.
Step 2: Created an empty embedding matrix.
Step 3: For each word in the cleaned text dictionary, we checked whether it existed in the pre-trained embeddings_index dictionary. If it existed, it added the corresponding embedding vector to the embedding matrix.
Step 4: If it did not exist, the row for that word was left as all zeros in the embedding matrix. The words that were not found in the pre-trained embeddings were added to the words_not_found list.
Step 5: Embedding matrix was ready and we passed it to the model.

**4.Test Train Validation split**
Input –> Feature extracted dataset from part 2
Output–> Feature extracted test,train and validation data

Step 1: We had done the test,train split in 60:40 ratio over both 'Cleaned Tweets' and 'Label' and saved them in respective variable

Step 2: Test,Train and Validation split was complete

**5.Define the model**
Input –>None
Output–> A sequential model with LSTM layer

Step 1: We import Sequential() from keras and saved it in a variable named as model.
Step 2: We added an Embedding layer at the first with parameters MAX_NB_WORDS, EMBEDDING_DIM that we had defined in part-2 .
Step 3: We had then added a special drop out layer with rate = 0.2 to reduce overfitting.
Step 4: Now we added a LSTM layer with 100 memory units with dropout rate of 0.2 for the input layer, and recurrent dropout rate = 0.2. This was to learn the to learn the temporal dependencies between words in the sequences.
Step 5: Now we added a dense() layer with 'n' output units and the activation function as 'softmax' or 'sigmoid'(according to binary or multiclass classification) which would gave the probability distribution over 'n' classes .
Step 6: Now our model was ready to compile.
Step 7: Now we compiled the model with categorical cross-entropy loss or binary cross-entropy, Adam optimizer, and accuracy metric for evaluation during training.

**6.Fit the model**
Input –>Pre-processed dataset from part 4 and compiled model from part 5.
Output–>Training of LSTM model using the preprocessed dataset and a trained model.

Step 1: We had fitted our model with the preprocessed dataset and predefined
Epoch number and batch size as our choice.
Step 2: Made the validation split of 50 % over test data.
Step 3: Our Model was trained.

7.Check the accuracy of the model
Step 1: Using the trained model, we predicted the labels of each tweet from test data set and store them in a variable.
Step 2: From sklearn.metrics we have imported the predefined objects and using the output from step 1 and test data from part 3 we have got our Overall accuracy , F1 score and confusion matrix.
Step 3: Record accuracy, F1 score and confusion matrix in a word document.

***Algorithm for BiLSTM Model (Without FASTTEXT Embedding):***

**1.Data Preprocessing**
Input: .csv file containing the labelled tweets
Output: .csv file containing the cleaned tweets with their corresponding label

Step 1: Read the tweets of the dataset.
Step 2: Remove the username starting with @.
Step 3: Remove the url from each single tweet.
Step 4: Replace multiple white spaces with single white space.
Step 5: Preprocessed dataset is ready.

**2.Feature Extraction**
Input: Preprocessed Dataset
Output: Feature extracted and moved for test train split

Step 1: The maximum sequence length was set to 250 as an initialization step.
Step 2: The embedding dimension was defined as 100.
Step 3: From keras.preprocessing we import Tokenizer and apply it on the column of ‘Cleaned Tweets’.
Step 4: The texts_to_sequences function from keras.preprocessing was imported and applied to the 'Cleaned
Tweets’ column ,with the resulting output saved in a variable.
Step 5: The pad_sequences function from keras.preprocessing was imported and applied to the variable obtained in step 4, using the padding_length value initialized in step 1 as a parameter.
Step 6 : The column associated with the tweet labels was one-hot encoded.
Step 7: With these steps completed, the feature extraction process was finished, resulting in the extracted
features ready for further usage and the subsequent test-train split.

**3.Test Train Validation split**
Input: Preprocessed Dataset
Output: Feature extracted and moved for test train split

Step 1: The dataset was divided into test and train sets in a 60:40 ratio, encompassing both the 'Cleaned Tweets' and 'Label' columns. These splits were saved in their respective variables.
Step 2: Test train validation is completed.

**4.Define the model**
Input: None
Output: A BiLSTM layer sequential model

Step 1: Sequential() from keras was imported, and the model was initialized and saved as a variable named "model".
Step 2: An Embedding layer was added to the model as the first layer, using the parameters MAX_NB_WORDS and EMBEDDING_DIM defined in part-2.
Step 3: A dropout layer was introduced to reduce overfitting, with a dropout rate of 0.2.
Step 4: A dense() layer was added to produce 'n' output units, and the activation function was set to 'softmax' or 'sigmoid' based on binary or multiclass classification requirements. This layer provided a probability distribution over the 'n' classes.
Step 5 : With the model architecture in place, it was ready for compilation.
Step 6: The model was compiled with categorical cross-entropy loss (for multiclass classification) or binary cross-entropy loss (for binary classification), Adam optimizer, and accuracy metric for evaluating the model's performance during training.

**5.Fit the model**
Input: Pre-processed dataset from part 3 and compiled model from part 4 .
Output: Training of BiLSTM model using the preprocessed dataset and a trained model .

Step 1: The model was fitted with the preprocessed dataset, utilizing the specified number of epochs and batch size. This involved feeding the training data into the model and iteratively updating the model's parameters to minimize the loss and improve performance.
Step 2: A validation split was performed, allocating 50% of the test data for validation purposes.
Step 3: The model is trained

***Algorithm for BiLSTM Model (With FASTTEXT Embedding):***

**1.Data Preprocessing**
Input: .csv file containing the labelled tweets
Output: .csv file containing the cleaned tweets with their corresponding label

Step 1: Read the tweets of the dataset.
Step 2: Remove the username starting with @.
Step 3: Remove the url from each single tweet.
Step 4: Replace multiple white spaces with single white space.
Step 5: Preprocessed dataset is ready.

2.Load fasttext model
Input: None
Output: Load the pretrained bin file from fasttext model.

Step 1: Download the pretrained bin file from the fasttext file.
Step 2: Unzip the downloaded file and extract the bin file

**3. Feature Extraction**
Input: Preprocessed Dataset
Output: Feature extracted and moved for test train split
The process can be divided into two parts, involving the manipulation of raw text tweets and the creation of an embedding matrix for these words. Here is a breakdown of the steps
Step 1: we imported the Tokenizer function from keras.preprocessing and applied it to the 'Cleaned Tweets' column. The resulting tokenized representation of the tweets was saved in a variable.
Step 2: we imported the texts_to_sequences function from keras.preprocessing and applied it to the tokenized tweets obtained in the previous step. This conversion process transformed the tokenized tweets into sequences, which were saved in a variable.
Step 3: we imported the pad_sequences function from keras.preprocessing and applied it to the sequences obtained in the previous step. This step involved padding the sequences with a specified padding length, which was initialized in step 1. The padded sequences were then stored in a variable.
Step 4: we performed one-hot encoding on the column associated with the labels of the tweets. This process converted the labels into a binary representation that could be easily processed by the model.
Step 5 : With these steps completed, the feature extraction process reached its conclusion

Now we move onto the next part
Step 1: We initialize the embedding dimension, determining the size of each embedding vector.
Step 2: An empty embedding matrix is created, serving as a container for the embedding vectors.
Step 3: We generate the vector for each word which we get after tokenization from the previous part, by the help of the .bin file.
Step 4: We add each vector into the embedding matrix
Step 5: In the event that a word is not found in the pre-trained embeddings, the respective row in the embedding matrix is filled with zeros, indicating the absence of an embedding vector for that word. These words are recorded in the words_not_found list for reference.
Step 6: This matrix can now be passed to the model as an input.

**4.Test Train Validation Split**
Input: Embeddding matrix from previous part
Output: A BiLSTM layer sequential model

Step 1: we import Sequential() from keras and instantiate a variable named model to represent our sequential model.
Step 2: The model is then configured with an Embedding layer, where we specify the parameters nb_words, embed_dim, and input_length. The weights parameter is utilized to assign the pre-trained embedding matrix, which remains non-trainable in this case.
Step 3: To address the issue of overfitting, we incorporate a dropout layer into the model, setting the dropout rate to 0.2. This layer aids in reducing the likelihood of overfitting by randomly dropping out a fraction of the input units during training.
Step 4: An BiLSTM layer is added, consisting of 100 memory units. The dropout rate for the input layer is set to 0.2, while the recurrent dropout rate is also set to 0.2. This layer enables the model to learn the temporal dependencies between words in the input sequences.
Step 5: A dense layer is appended to the model, with 'n' output units and an activation function of either 'softmax' or 'sigmoid,' depending on whether the classification task is binary or multiclass. This layer provides a probability distribution over the 'n' classes, facilitating the classification process.
Step 6: With the model's structure established, it is now ready to be compiled
Step 7: During the compilation step, the model is assigned the appropriate loss function—categorical cross-entropy for multiclass classification or binary cross-entropy for binary classification. Additionally, an Adam optimizer is utilized to optimize the model's performance, and the accuracy metric is employed for evaluation during the training process.

**5.Fit the Model**
Input: Pre-processed dataset from part 3 and compiled model from part 4 .
Output: Training of BiLSTM model using the preprocessed dataset and a trained model .

Step 1: Our model has been trained by fitting it with the preprocessed dataset, using a predefined number of epochs and batch size according to our specifications
Step 2: A validation split was performed, allocating 50% of the test data for validation purposes.
Step 3: The model is trained

**6.Check the accuracy of the model**
Input: Trained model from part 5 and test dataset from part 3
Output: Overall accuracy of the model.
Step 1: After training our model we proceeded to predict the labels of each tweet in the test dataset using the trained model. The predicted labels were stored in a variable for further analysis
Step 2: We imported the necessary predefined objects from the sklearn.metrics module. By comparing the predicted labels with the actual labels from part 3 of the test data, we obtained the overall accuracy, F1 score, and confusion matrix.
Step 3: The recorded accuracy, F1 score, and confusion matrix were documented in a Word document, ensuring that the evaluation results were properly recorded for future reference or analysis.

## 4. DATASET AND BASELINES

### 4.1 Dataset

The English, German and Italy dataset used here is collected from the authors of the paper "A Multilingual Evaluation for Online Hate Speech Detection": Michele Corazza, Stefano Menini Elena Cabrio, Sara Tonelli, Serena Villata.[10]

**English Dataset:** The English dataset was taken from the paper Hateful Symbols or Hateful People? Predictive Features for Hate Speech Detection on Twitter [35]. It contains 15,777 tweets and classified over three different classes. 10841 tweets are labelled as "non hate", while 3017 are labelled as "racism" and 1919 are labelled as "sexism".

**Italian Dataset:** The Italian dataset was taken from the paper Overview of the EVALITA 2018 Hate Speech Detection Task [37]. It contains 3000 tweets and classified over two different classes. The distribution of tweets among these two classes is as followed. 2028 tweets are labelled as "hate" and 972 are labelled as "non hate".

**German Dataset:** The German dataset is taken from the paper Overview of the GermEval 2018 Shared Task on the Identification of Offensive Language [38]. It contains 3031 tweets and classified over two different classes. 2061 tweets are labelled as "non hate" while 970 are labelled as "hate".

**Bengali Dataset:** The Bengali dataset was taken from the paper DeepHateExplainer: Explainable Hate Speech Detection in Under-resourced Bengali Language [36]. It contains 3419 tweets and classified over five different classes. 1379 tweets are labelled as "geopolitical", 629 are labelled as "personal", 592 are labelled as "political", 502 are labelled as "religious", and 316 are labelled as "abusive".

### 4.2 Baseline

A robust foundation for the evaluation of our proposed approach is established by referencing a pivotal work titled "A Multilingual Evaluation for Online Hate Speech Detection" [10]. Authored by Michele Corazza, Stefano Menini, Elena Cabrio, Sara Tonelli, and Serena Villata, this study serves as the baseline against which the efficacy of our approach is measured.

In this paper, a robust neural architecture is presented, demonstrating satisfactory performance across different languages: English, Italian, and German. An extensive analysis of experimental results is conducted, delving into the contributions of various components, both from the architectural perspective (i.e., Long Short Term Memory, Gated Recurrent Unit, and bidirectional Long Short Term Memory) and the feature selection standpoint (i.e., ngrams, social network-specific features, emotion lexica, emojis, word embeddings). This in-depth analysis employs three freely available datasets for hate speech detection on social media in English, Italian, and German.

## 5. EVALUATION AND COMPARISON OF RESULTS

A multilingual dimension is introduced to the evaluation through datasets in English, German, Italian, and Bengali, thereby reflecting the adaptability of the models to different linguistic contexts. In this study, not only is the performance of these models in hate speech detection assessed, but the results are also compared against a baseline model, serving as a benchmark for evaluating the advancements made by these sophisticated models.

**Table 1**
Hate speech Detection: English Dataset

| Metric | Class | LSTM | | BiLSTM | | BERT |
|---|---|---|---|---|---|---|
| | | Without FASTTEXT | With FASTTEXT | Without FASTTEXT | With FASTTEXT | |
| Precision | None | 0.85 | 0.84 | 0.88 | 0.84 | 0.93 |
| | Sexism | 0.72 | 0.77 | 0.69 | 0.79 | 0.90 |
| | Racism | 0.74 | 0.73 | 0.77 | 0.73 | 0.95 |
| Recall | None | 0.90 | 0.91 | 0.88 | 0.91 | 0.97 |
| | Sexism | 0.66 | 0.54 | 0.76 | 0.53 | 0.87 |
| | Racism | 0.56 | 0.71 | 0.65 | 0.71 | 0.82 |
| F1 Score | None | 0.87 | 0.88 | 0.88 | 0.88 | 0.95 |
| | Sexism | 0.69 | 0.64 | 0.72 | 0.64 | 0.89 |
| | Racism | 0.64 | 0.72 | 0.77 | 0.72 | 0.88 |

**Table 2**
Hate speech Detection: Italian Dataset

| Metric | Class | LSTM | | BiLSTM | | BERT |
|---|---|---|---|---|---|---|
| | | Without FASTTEXT | With FASTTEXT | Without FASTTEXT | With FASTTEXT | |
| Precision | Non-Hate | 0.81 | 0.84 | 0.82 | 0.82 | 0.68 |
| | Hate | 0.68 | 0.77 | 0.68 | 0.67 | 0.81 |
| Recall | Non-Hate | 0.85 | 0.81 | 0.87 | 0.85 | 0.58 |
| | Hate | 0.72 | 0.57 | 0.62 | 0.62 | 0.86 |
| F1 Score | Non-Hate | 0.83 | 0.83 | 0.84 | 0.84 | 0.63 |
| | Hate | 0.63 | 0.63 | 0.65 | 0.64 | 0.84 |

**Table 3**
Hate speech Detection: German Dataset

| Metric | Class | LSTM | | BiLSTM | | BERT |
|---|---|---|---|---|---|---|
| | | Without FASTTEXT | With FASTTEXT | Without FASTTEXT | With FASTTEXT | |
| Precision | Non-Hate | 0.78 | 0.75 | 0.76 | 0.76 | 0.91 |
| | Hate | 0.54 | 0.51 | 0.56 | 0.53 | 0.91 |
| Recall | Non-Hate | 0.80 | 0.82 | 0.84 | 0.76 | 0.80 |
| | Hate | 0.50 | 0.40 | 0.43 | 0.46 | 0.96 |
| F1 Score | Non-Hate | 0.79 | 0.78 | 0.80 | 0.79 | 0.85 |
| | Hate | 0.52 | 0.45 | 0.48 | 0.49 | 0.94 |

**Table 4**
Hate speech Detection: Bengali Dataset

| Metric | Class | LSTM | | BiLSTM | | BERT |
|---|---|---|---|---|---|---|
| | | Without FASTTEXT | With FASTTEXT | Without FASTTEXT | With FASTTEXT | |
| Precision | Gender abusive | 0.80 | 0.80 | 0.57 | 0.83 | 0.69 |
| | Geopolitical | 0.70 | 0.67 | 0.68 | 0.56 | 0.94 |
| | Personal | 0.64 | 0.57 | 0.58 | 0.60 | 0.90 |
| | Political | 0.67 | 0.63 | 0.65 | 0.62 | 0.87 |
| | Religious | 0.78 | 0.63 | 0.79 | 0.67 | 0.91 |
| Recall | Gender abusive | 0.85 | 0.82 | 0.87 | 0.79 | 0.76 |
| | Geopolitical | 0.47 | 0.37 | 0.45 | 0.40 | 0.96 |
| | Personal | 0.38 | 0.46 | 0.53 | 0.41 | 0.73 |
| | Political | 0.19 | 0.32 | 0.24 | 0.32 | 0.96 |
| | Religious | 0.55 | 0.71 | 0.60 | 0.62 | 0.91 |
| F1 Score | Gender abusive | 0.83 | 0.81 | 0.81 | 0.81 | 0.72 |
| | Geopolitical | 0.57 | 0.47 | 0.54 | 0.46 | 0.95 |
| | Personal | 0.48 | 0.51 | 0.55 | 0.49 | 0.81 |
| | Political | 0.30 | 0.43 | 0.35 | 0.42 | 0.91 |
| | Religious | 0.64 | 0.67 | 0.68 | 0.63 | 0.91 |

**Table 6**
Comparative Analysis of Hate Speech Detection Results Across Datasets with Baseline Model

| Dataset | Metric | LSTM | | BiLSTM | | BERT | Baseline |
|---|---|---|---|---|---|---|---|
| | | Without FASTTEXT | With FASTTEXT | Without FASTTEXT | With FASTTEXT | | |
| English | Average Precision | 0.77 | 0.78 | 0.78 | 0.79 | 0.93 | 0.820 |
| | Average Recall | 0.71 | 0.72 | 0.76 | 0.72 | 0.89 | 0.825 |
| | Average F1 Score | 0.73 | 0.75 | 0.79 | 0.75 | 0.91 | 0.823 |
| Italian | Average Precision | 0.75 | 0.73 | 0.75 | 0.66 | 0.75 | 0.803 |
| | Average Recall | 0.79 | 0.69 | 0.75 | 0.64 | 0.72 | 0.806 |
| | Average F1 Score | 0.73 | 0.73 | 0.75 | 0.64 | 0.74 | 0.805 |
| German | Average Precision | 0.66 | 0.63 | 0.75 | 0.65 | 0.91 | 0.754 |
| | Average Recall | 0.65 | 0.61 | 0.74 | 0.61 | 0.88 | 0.762 |
| | Average F1 Score | 0.66 | 0.62 | 0.74 | 0.64 | 0.90 | 0.758 |

## 6. CONCLUSION

This work primarily focuses on the detection of hate speech in social media messages across multiple languages, including Bengali, English, German, and Italian.
Our objective was to identify a reliable and high-performing neural network for this task. In addition, we conducted a comprehensive evaluation of various components that are commonly used in hate speech detection, such as different types of embeddings, the inclusion of additional text-based features, the normalization of hashtags, and the influence of emojis.
By examining the performance and contribution of these different components, we aimed to gain insights into their effectiveness in hate speech detection across various languages. This research endeavor sheds light on the importance of these factors and provides valuable guidance for developing robust hate speech detection models.
After robust experimenting using both methods and recording the results, we came to the conclusion that LSTM, BiLSTM and BERT are proper neural network models for this type of work.
It was also proven that giving the inputs in vector form increased the accuracy of the model for some cases. So fasttext method was also helpful to increase the accuracy of this model.
The availability of large, labelled hate speech datasets is limited. This scarcity hinders the training of accurate and reliable BiLSTM network, as the network heavily rely on diverse and representative data.
Hate speech detection online is a complex problem that demands attention. Despite our efforts, achieving 100% accuracy remains elusive due to various factors, including system limitations like limited RAM and GPU resources.
Sentences with complex syntactic structures (eg, containing multiple negations or interrogative sentences) are common, both in false-positive and false-negative sentences.

## 7. FUTURE WORKS

Future research in multilingual hate speech detection should focus on addressing the identified challenges and advancing the capabilities of these models. Alternative RNN models beyond LSTM and BiLSTM could be explored to enhance performance. Leveraging language-specific fasttext models and experimenting with various embeddings may mitigate issues related to slang, masked hate words, and challenges in discerning sarcasm. Developing larger, diverse labeled datasets is imperative to improve training accuracy, particularly for less-represented languages. Hybrid models and advanced techniques could be investigated to combine the strengths of different neural network architectures. Additionally, efforts should be directed towards improving model interpretability and transparency, considering the ethical implications of biases in hate speech detection. This research lays the groundwork for future studies to refine and innovate hate speech detection systems across diverse linguistic and contextual landscapes.